\documentclass[runningheads]{llncs}
\usepackage[T1]{fontenc}
\usepackage{graphicx}
\usepackage{booktabs}
\usepackage{multirow}
\usepackage[hidelinks]{hyperref}
\usepackage[dvipsnames]{xcolor}
\newcommand{\etal}{\textit{et al.~}}

\newcommand*\samethanks[1][\value{footnote}]{\footnotemark[#1]}

\begin{document}
\title{SafeTriage: Facial Video De-identification for Privacy-Preserving Stroke Triage}
\titlerunning{SafeTriage: Privacy-Preserving Stroke Triage}

\author{Tongan Cai\inst{1}\thanks{These authors contributed equally to this work.} 
\and Haomiao Ni\inst{2}\samethanks
\and Wenchao Ma\inst{1}\samethanks
\and Yuan Xue\inst{3} 
\and Qian Ma\inst{1} \and 
\\Rachel Leicht\inst{4} \and Kelvin Wong\inst{4} \and John Volpi\inst{4} \and Stephen T.C. Wong\inst{4} \and 
\\James Z. Wang\inst{1} \and Sharon X. Huang\inst{1}}
%index{Cai, Tongan}
%index{Ni, Haomiao}
%index{Ma, Wenchao}
%index{Xue, Yuan}
%index{Ma, Qian}
%index{Leicht, Rachel}
%index{Wong, Kelvin}
%index{Volpi, John}
%index{Wong, Stephen}
%index{Wang, James}
%index{Huang, Sharon}
\authorrunning{Cai et al.}
\institute{The Pennsylvania State University, University Park, PA, USA \and 
The University of Memphis, Memphis, TN, USA
\and
The Ohio State University, Columbus, OH, USA
\and 
Houston Methodist Hospital, Houston, TX, USA
}

\let\oldmaketitle\maketitle
\renewcommand{\maketitle}{\oldmaketitle\setcounter{footnote}{0}}
\maketitle              % typeset the header of the contribution
\begin{abstract}
Effective stroke triage in emergency settings often relies on clinicians' ability to identify subtle abnormalities in facial muscle coordination. While recent AI models have shown promise in detecting such patterns from patient facial videos, their reliance on real patient data raises significant ethical and privacy challenges---especially when training robust and generalizable models across institutions. To address these concerns, we propose \textit{SafeTriage}, a novel method designed to de-identify patient facial videos while preserving essential motion cues crucial for stroke diagnosis. SafeTriage leverages a pretrained video motion transfer (VMT) model to map the motion characteristics of real patient faces onto synthetic identities. This approach retains diagnostically relevant facial dynamics without revealing the patients' identities. To mitigate the distribution shift between normal population pre-training videos and patient population test videos, we introduce a conditional generative model for visual prompt tuning, which adapts the input space of the VMT model to ensure accurate motion transfer without needing to fine-tune the VMT model backbone. Comprehensive evaluation, including quantitative metrics and clinical expert assessments, demonstrates that SafeTriage-produced synthetic videos effectively preserve stroke-relevant facial patterns, enabling reliable AI-based triage. Our evaluations also show that SafeTriage provides robust privacy protection while maintaining diagnostic accuracy, offering a secure and ethically sound foundation for data sharing and AI-driven clinical analysis in neurological disorders.
\end{abstract}

\section{Introduction}
Stroke is a leading cause of disability and mortality worldwide \cite{johnson2016stroke}. Timely detection and intervention significantly enhance survival rates and the quality of life for stroke patients. In emergency room (ER) triage, potential strokes are often identified by observing subtle abnormalities in oral-facial movements. However, the shortage of experienced neurologists \cite{leira2013growing} and the nuanced nature of these indicators \cite{yu2020toward} undermine the reliability of triage in critical stroke situations. 

\begin{figure}
    \centering
    \includegraphics[width=0.70\linewidth]{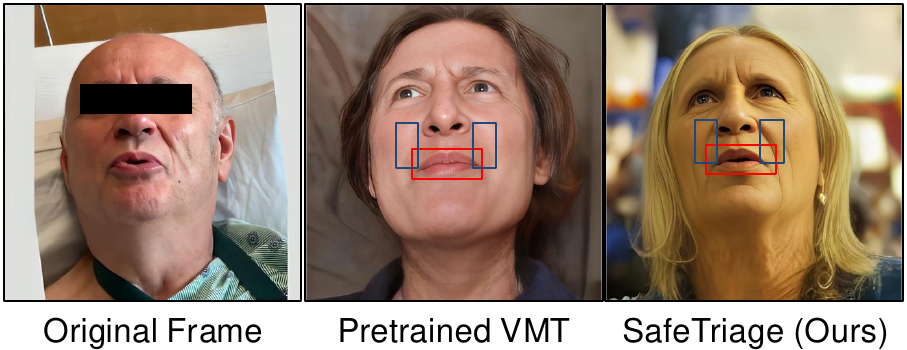}
    \caption{Illustration of the advantages of the SafeTriage framework. A direct application of a pretrained video motion transfer (VMT) model, trained with normal population faces, fails to accurately capture subtle motions (e.g., lip movements highlighted by the \textcolor{red}{red} box) and facial asymmetries (e.g., asymmetries highlighted by the \textcolor{blue}{blue} boxes) in patient faces. In contrast, SafeTriage overcomes these limitations through visual prompt tuning (VPT) to effectively adapt the pretrained VMT model for patient faces.}
    \label{fig:teaser}
\end{figure}

Recent advances in machine intelligence have shown promise in identifying neurological disorders from visual and audio inputs. For example, Cai \etal \cite{cai2022deepstroke} introduced \textit{DeepStroke}, an effective stroke triage framework tailored for ER environments, utilizing video and speech data for stroke detection. Zhuang \etal \cite{zhuang2021video} developed a method for evaluating facial weakness from videos. Despite the progress, the success of existing stroke triage methods and AI-assisted models heavily depends on access to datasets of real patient videos. Privacy and ethical considerations, however, severely limit sharing these sensitive data across institutions. As a result, models are often trained on limited, in-house datasets, hindering their robustness, scalability, and generalizability. 

In the broader domain of video analysis, various strategies have been explored for privacy protection and de-identification. Dave \etal \cite{dave2022spact} introduced a self-supervised framework for action recognition that employs a minimax optimization strategy. This framework aims to minimize the cost of action recognition while maximizing privacy through a contrastive self-supervised loss. Although it effectively removes privacy attributes, the actions in the generated videos are only recognizable by the system's modules and not by humans. Xia \etal \cite{xia2023diffslva} developed a text-guided sign language video anonymization technique using a large-scale diffusion model \cite{rombach2022high}, which relies on text inputs for guidance---inputs that are often unavailable in patient facial videos. Meanwhile, approaches in medical imaging---such as using Generative Adversarial Networks (GAN)~\cite{goodfellow2014generative} for de-identification of skin lesion images~\cite{Bissoto_2021_CVPR}, MRI results~\cite{shin2018medical}, and Chest X-Rays~\cite{montenegro2024anonymizing}---have seen some success. Translating these strategies to clinical videos is less straightforward. Hou \etal \cite{hou2023artificial} utilized landmark-guided face morphing on seizure patient videos, and Flouty \etal \cite{flouty2018faceoff} applied facial blurring in surgical settings. While these methods obscure subject identities, they fail to preserve the critical facial motions necessary for video-based facial analysis, thus hindering large-scale data sharing essential for advancing AI-assisted medical diagnosis and intervention. Zhu \etal \cite{zhu2024facemotionpreserve} recently proposed a generative approach for facial de-identification and medical information preservation, but their method relies on a conditional GAN model trained from scratch and uses faces of healthy volunteers, which may limit its scalability and adaptability to general applications.

To address the persistent challenges of data scarcity and to facilitate the sharing of patient facial videos, we present \textit{SafeTriage}, a novel framework that de-identifies patient facial videos by retargeting them onto synthetic appearances while preserving the critical facial motion patterns needed for accurate stroke diagnosis. SafeTriage leverages a pretrained video motion transfer (VMT) model to map patient-specific motions onto synthetic faces, generating videos that combine synthetic appearances with pathological motion dynamics. Directly applying the VMT model to patient data, however, introduces a significant domain gap, as the model is originally trained on videos of healthy individuals (see Fig.~\ref{fig:teaser}).
To mitigate this, we draw inspiration from recent advancements in visual prompt tuning (VPT) \cite{jia2022visual} and introduce a conditional generation module that adapts the input space of the VMT model. This module reduces distributional shifts between training and testing while keeping the model backbone frozen. Unlike traditional model fine-tuning, our approach requires training only a small number of parameters in the introduced generative model, thereby largely preserving the prior knowledge embedded in the pretrained VMT model. Comprehensive experiments and evaluation demonstrate that SafeTriage can generate high-quality, de-identified synthetic videos that effectively retain stroke-relevant diagnostic features, achieving an optimal balance between privacy protection and diagnostic accuracy.

\begin{figure}[t]
    \centering
\includegraphics[width=\linewidth]{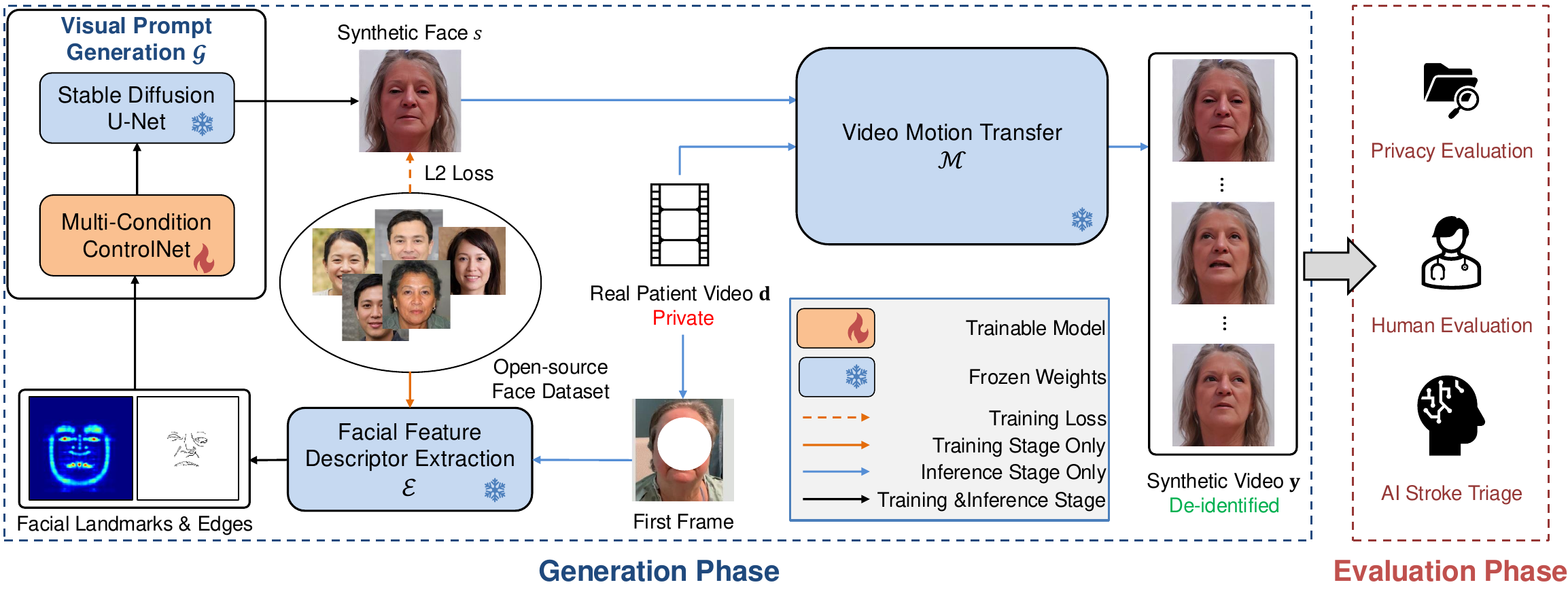}
    \caption{Overview of the proposed SafeTriage framework, which comprises two main phases: the generation phase and the evaluation phase. The generation phase is based on a \textit{frozen} video motion transfer model $\mathcal{M}$. Given a private video of a real patient $\mathbf{d}$, SafeTriage introduces a visual prompt generation model, $\mathcal{G}$, to generate a synthetic face, $s$, with facial asymmetries and head pose aligned to the first frame of $\mathbf{d}$. Then $\mathcal{M}$ is employed to generate a de-identified synthetic video, $\mathbf{y}$, given the conditionally-generated input $s$ and real patient video $\mathbf{d}$. In the evaluation phase, the diagnostic utility and privacy preservation of the synthetic video $\mathbf{y}$ are assessed. It is important to note that the training of $\mathcal{G}$ only uses publicly available, non-private face data.}
    \label{fig:framework}
\end{figure}

\section{Methodology}
Figure~\ref{fig:framework} presents an overview of the proposed SafeTriage framework for privacy-preserving stroke triage. The framework comprises two phases: a generation phase and an evaluation phase. In the generation phase, de-identified synthetic videos are generated while preserving the diagnostic motion characteristics of real patient videos. This is achieved through two main components: (1) visual prompt generation and (2) video motion transfer (VMT). To address the limited availability of patient video data, we leverage a large-scale pretrained VMT model as the foundation model of our framework. To minimize computational cost and avoid using privacy-sensitive patient data for model training, rather than fine-tuning the VMT network using patient data, we introduce a conditional generation module to produce a visual prompt (i.e., a synthetic subject image) that has the subject head pose and facial edge features aligned to the first frame of a driving patient video. Using the conditionally generated subject images can effectively help the VMT model adapt to a new patient distribution while keeping its backbone parameters frozen. 
In the evaluation phase, we assess the synthetic videos in three aspects: (1) human evaluation to ascertain visual realism and the preservation of diagnostic patterns; (2) privacy evaluation to measure the effectiveness of de-identification; and (3) performance assessment of an AI-based stroke triage model with synthetic videos to verify that diagnostic accuracy is maintained. 
Details are presented in the following sections.

\subsection{The Generation Phase}
\label{sec:model_gen}

\begin{figure}[t]
    \centering \includegraphics[width=0.8\linewidth]{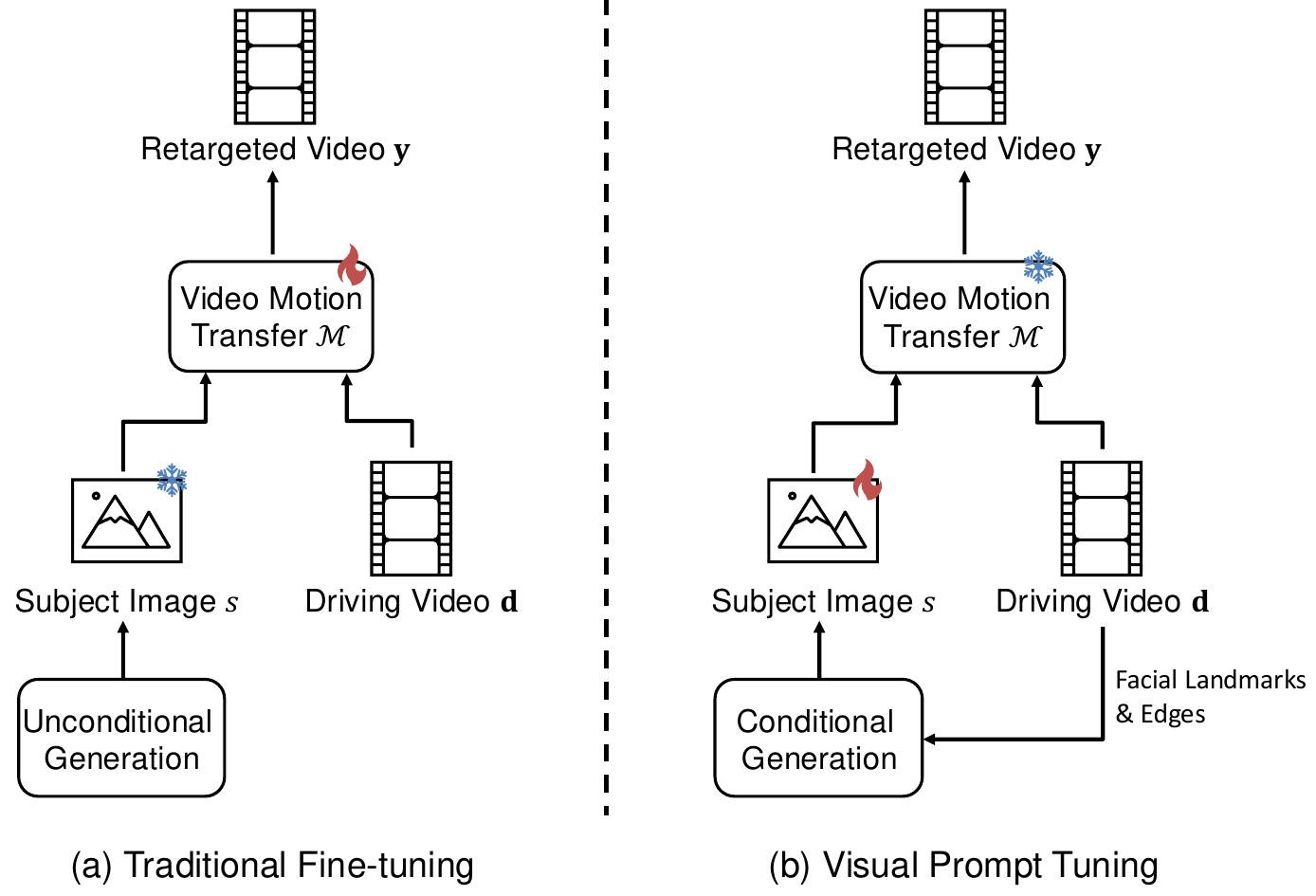}
    \caption{Illustration of Visual Prompt Tuning. Given a subject image $s$ and a driving video $\mathbf{d}$, the video motion transfer model $\mathcal{M}$ generates a retargeted video $\mathbf{y}$ with the appearance of the subject image and motion in the driving video.  Similar to text prompts in large language models (LLMs), $s$ and $\mathbf{d}$ can be viewed as \textit{visual} input prompts for the model $\mathcal{M}$. To adapt $\mathcal{M}$ to a new distribution, unlike traditional methods that fine-tune the model backbone, we employ an efficient and effective visual prompt engineering technique that uses a conditional generation process to generate the visual prompt $s$ that is specifically \textit{tuned} to be aligned with the first frame of the driving video 
$\mathbf{d}$, enabling $\mathcal{M}$ to perform high-quality transfer of out-of-distribution motion. 
}
    \label{fig:vpt}
\end{figure}

\hspace{\parindent}\textbf{Visual Prompt Generation.} Considering the limited availability of patient videos and the privacy concerns associated with using such data for model training, our framework leverages a VMT foundation model, $\mathcal{M}$, pretrained on publicly available, large-scale datasets of in-the-wild facial videos of primarily healthy individuals. However, directly applying this pretrained VMT model to patient video motion transfer results in degraded performance due to the distributional shift between healthy and patient cohorts (see Fig.~\ref{fig:teaser}). Traditional parameter fine-tuning methods, such as full fine-tuning, partial fine-tuning (e.g., updating a subset of parameters \cite{zaken2021bitfit}), and additive fine-tuning (e.g., using external modules like adapters \cite{chen2024conv,chen2022adaptformer,houlsby2019parameter}), can address domain gaps but often incur notable computational costs and, in the context of patient facial videos, pose an additional risk of privacy leakage. To minimize training overhead and the risk of privacy leakage, we adopt an approach inspired by visual prompt tuning (VPT)~\cite{jia2022visual}. Similar to text prompts in large language models (LLMs), the input subject image and driving video for the VMT foundation model can be viewed as visual prompts. The subject image is typically a face synthesized by a generative AI model with a pseudo-identity (i.e., a non-real identity), which will be later used to conceal the real identity of patients. As shown in Fig.~\ref{fig:vpt}, our proposed VPT approach differs from the traditional fine-tuning method by freezing the VMT model backbone while using a conditionally generated subject image prompt specifically tuned to align with privacy-safe features of the driving patient video.

To achieve this, we employ a conditional facial generative model $\mathcal{G}$ to synthesize a pseudo-identity face image $s$ that has its head pose and facial features (such as facial asymmetries) aligned with the real patient video $d$. Specifically, a pretrained facial feature descriptor model $\mathcal{E}$ is used to extract facial edges $e$ and landmarks $m$ from the first frame of the patient video $d$. The extracted facial edges are mainly in the eye, mouth and cheek regions to capture potential pathological features such as asymmetries (see the edge feature example shown in Fig. \ref{fig:framework}). 
The facial edge and landmark features serve as conditions for $\mathcal{G}$ to generate the visual prompt $s$, which also ensures alignment of the head pose between $s$ and the first frame of $d$.
To maintain the quality of the generated visual prompts, we filter out low-quality synthetic images using an off-the-shelf image quality assessment tool. 
As a general framework, we can use various model backbones to implement the model $\mathcal{G}$. Here we choose ControlNet \cite{zhang2023adding} for its high performance and efficiency. To avoid identity leakage, we train the model $\mathcal{G}$ exclusively with publicly available face data, ensuring that no private patient data is used during the training process. More details can be found in Section ~\ref{sec:model_implement}.

\textbf{Video Motion Transfer.} Let $s$ be a pseudo-identity face image, referred to as the subject image. Let $\mathbf{d}=\langle d_1, d_2, \cdots, d_N\rangle$ be a video of a patient speaking, such as describing the ``cookie theft'' picture commonly used for rating aphasia severity in the NIH Stroke Scale. $\mathbf{d}$, consisting of $N$ individual frames $d_i, i=1,\ldots,N$, is used as the driving video for motion transfer from the patient to the pseudo-identity subject. 
We utilize a pretrained VMT model $\mathcal{M}$ to generate an output video $\mathbf{y}=\langle y_1, y_2, \cdots, y_N\rangle$, where the identity in $y_i$'s is inherited from the synthetic face $s$ and the motions are derived from the original patient video $d_i$'s. A typical warp-based VMT network~\cite{ni20243d,siarohin2019first} consists of three main stages: (1) semantic correspondence learning, (2) deformation flow modeling, and (3) warping and inpainting, with inpainting often handled by a generator. 
Specifically, the VMT model $\mathcal{M}$ first identifies semantic correspondences (i.e., locations in multiple images that share the same semantic meaning \cite{hedlin2024unsupervised}) between $s$ and $d_i$ by extracting keypoints (either 2D \cite{siarohin2019first} or 3D \cite{wang2021one}) or regions \cite{siarohin2021motion} in the feature space. Using these correspondences, $\mathcal{M}$ predicts the dense deformation flow, aligning the pose of the subject in $s$ with the pose in $d_i$ via a warp operation. Due to potential occlusions, deformation flow alone may not be sufficient for generation as occluded parts cannot be recovered by image warping. Therefore, a generator to inpaint occluded areas is employed in $\mathcal{M}$ to produce the final complete and realistic output. This motion transfer process ensures the de-identification of patient videos while preserving the essential motion patterns for downstream tasks. For implementation, various VMT model backbones can be used for motion transfer; we discuss our specific choice in Section \ref{sec:model_implement}. 

\subsection{The Evaluation Phase} \label{sec:evaluation}
\hspace{\parindent}\textbf{Human Expert Evaluation.} 
We conduct a human evaluation to assess the \textit{visual realism} and \textit{diagnostic utility} of the generated videos $\mathbf{y}$. First, qualified visual content analysts evaluate the visual realism of each synthetic video by responding to the prompt: ``\textit{Do you think this video is realistic?}''. The analysts choose from the following options ``A. \textit{Very realistic},'' ``B. \textit{Moderately realistic},'' or ``C. \textit{Not realistic at all.}'' Only those synthetic videos unanimously rated as ``Very realistic'' proceed to a subsequent clinical review to validate the preservation of diagnostic patterns. During the clinical review, we perform a paired user study in which qualified clinicians compare each synthetic video generated by our model with its corresponding real patient video. The clinicians evaluate each pair of videos by answering the question: ``\textit{Do you think the diagnostic patterns between these two videos are consistent?}''. Clinicians respond with either ``Yes'' or ``No''.  

\textbf{Privacy Evaluation.} We validate the privacy-protection ability of the SafeTriage method by comparing the facial embedding vectors of real and synthetic videos generated with a pretrained face recognition network VGG-Face~\cite{parkhi2015deep}. The VGG-Face model achieves 98.9\% recognition accuracy on the commonly-used benchmark FaceNet~\cite{schroff2015facenet} while being computationally efficient. For each patient-to-synthetic retargeting case, we randomly select pairs of two frames and calculate the similarity between their facial embedding vectors. The two frames are either from the same real video of a patient, or one from the real video and the other from the corresponding synthetic video. We demonstrate the change in identity similarity score distribution before and after applying the SafeTriage framework.

\textbf{AI-Assisted Stroke Triage.} 
An additional quantitative evaluation is conducted by comparing stroke classification results using real patient videos vs. generated synthetic videos, by an AI-based stroke triage model. We adopt the 
\textit{DeepStroke}~\cite{cai2022deepstroke} model, which was developed for the task of stroke triage using patient facial videos. All the subjects used for training and testing the \textit{DeepStroke} model are patients visiting the ER and only some of them have a stroke confirmed by diffusion-weighted MRI. 
The \textit{DeepStroke} network backbone is a temporal-averaging ResNet-34 \cite{he2016deep} that treats all frames in one video as having the same label. The network uses the adjacent-frame difference vectors as input and performs a binary classification task (i.e., stroke vs. non-stroke) guided by a binary cross-entropy loss. The frame-level logits are stacked and averaged to generate the case-level prediction. 

To ensure robust evaluation and address the limited dataset size, we adopt the five-fold cross-validation approach to train and test the stroke triage model. Our experiments are designed to assess the preservation of diagnostic facial motion features by comparing the testing performance on real patient videos with that on their corresponding synthetic videos, using identical models trained on real data. Additionally, to explore the potential of our generated videos in facilitating data sharing and the development of future large AI models, we also examine the outcomes when synthetic videos are used for training and real or synthetic videos are used for validation within the cross-validation framework.

\section{Experiments}
\subsection{Dataset and Metrics} 
\hspace{\parindent}\textbf{Dataset.} We acquired a clinical dataset for this IRB-approved study from Houston Methodist Hospital. Patients suspected of stroke are selected during their ER visits without consideration of race or sex, ensuring equity and diversity. Patients are asked to perform two speech tasks while being video-recorded at a resolution of $1920\times1200$ at 30 frames per second (fps).
The videos are collected ``in the wild'' without imposing restrictions on the patients' positions, allowing for recordings of patients lying in bed, sitting, or standing, with varied backgrounds and lighting conditions.
Each video is paired with ground truth verification from diffusion-weighted MRI scans to determine stroke presence. Our study's cohort includes 113 patients, with 66 diagnosed with stroke and 47 identified as non-stroke but presenting other conditions based on MRI. 
On average, these videos have a length of 1,895 frames (about 63.16 seconds). 
Following prior work~\cite{cai2022deepstroke}, we construct a binary classification task, omitting stroke subtypes for simplicity. 

\textbf{Evaluation Metrics.} For the human evaluation described in Sec. \ref{sec:evaluation}, to quantitatively analyze the survey response results, we assign scores to the options provided: 3 points to option A, 2 points to option B, and 1 point to option C. Similarly, for responses regarding the preservation of diagnostic motion features in the generated videos, we assign 1 point for a ``Yes'' response and 0 point for a ``No'' response during the clinician review. For both evaluations, we calculate and report the Fleiss's kappa agreement score~\cite{fleiss1981measurement} to measure inter-rater agreement and the mean scores to provide an overall quantitative measure of performance. 

For the privacy protection evaluation, we use the cosine similarity (CSIM) \cite{zakharov2019few} between the aforementioned identity embedding vectors to measure identify distance and visualize the identity distance distributions for real-real video frame pairs and real-synthetic video frame pairs, respectively. 

For the evaluation of diagnostic utility with an AI stroke triage model, we use commonly referenced metrics, including accuracy, specificity, sensitivity, F1 score, and area under the ROC curve (AUC), to assess the effect of SafeTriage on the model’s stroke classification performance. We first establish the baseline of training the \textit{DeepStroke} triage model using real videos and testing on real videos. We then measure performance change from the baseline when testing the model on synthetic videos, or when both training and testing the model using synthetic videos. Furthermore, we measure the absolute change in prediction class probabilities and report the Mean Squared Error (MSE) between the prediction result (before applying softmax) from the real video baseline and the corresponding predictions from other train-test schemes involving synthetic videos.

\subsection{Model Implementation}
\label{sec:model_implement}
As mentioned in Section~\ref{sec:model_gen}, we use a conditional synthetic face generation model $\mathcal{G}$ for visual prompt tuning. 
Two types of facial feature descriptors, landmark heatmaps and edge maps, are used as the conditional inputs for model $\mathcal{G}$. The landmark heatmaps capture the pose of the patient in the driving video, and generating a synthetic face with matching pose helps the VMT model better retarget facial motions. 
Moreover, existing VMTs, pretrained primarily on normal faces, largely ignore pathological facial appearance features during the retargeting process---such as facial asymmetries and wrinkles (e.g., asymmetrically positioned eyes, eyebrows, or mouth corners)---which causes their direct application in transferring patient facial motion to produce subpar results with reduced clinical utility. To address this, we use edge maps as an additional condition in visual prompt tuning to retain such features. 

In practice, we first use \texttt{mmpose}\footnote{\url{https://github.com/open-mmlab/mmpose}} to detect facial landmarks and generate corresponding landmark heatmaps. 
For edges, since the edges of interest are around facial organs such as eyes, nose, and mouth, we utilize the detected landmarks to define bounding boxes around the interested facial organs and apply the Canny edge detector~\cite{john1986canny} within these bounding boxes to generate facial edge maps.

We implement the conditional generative model $\mathcal{G}$ with ControlNet~\cite{zhang2023adding}, a state-of-the-art computationally efficient model that reuses a frozen large pretrained text-to-image model, Stable Diffusion \cite{rombach2022high}, and introduces only a small number of trainable parameters to enable new conditioning controls.
To train this multi-condition ControlNet model, we collected 2,000 human face images from the publicly available FFHQ dataset~\cite{karras2019style}. We resized the images to a resolution of 256$\times$256 to accelerate training. The model was trained for 100 epochs with a batch size of 2, using two NVIDIA RTX 6000 GPUs. After training the model, multiple candidate synthetic faces are generated using the model given each set of conditions, and a public image quality assessment tool\footnote{\url{https://github.com/LAION-AI/aesthetic-predictor}} is applied to filter out low-quality generated images. A randomly chosen high-quality synthetic face is then further enhanced using CodeFormer \cite{zhou2022towards} and upsampled to 
512$\times$512 to be used as the visual prompt $s$.

We subsequently preprocess the original patient video to obtain the driving video $\mathbf{d}$ for motion retargeting by 1) eliminating constant camera roll, 2) square-cropping the face regions with sufficient borders, and 3) resizing to 512$\times$512. Given the subject image $s$ and the driving video $\mathbf{d}$, we implement the video motion transfer model $\mathcal{M}$ with LivePortrait \cite{guo2024liveportrait}, which achieves superior generalization due to large-scale training and advanced network architectures, while incorporating specialized modules for precise eye and lip retargeting. We use a specific version\footnote{\href{https://github.com/KwaiVGI/LivePortrait/tree/7fda929bf24e6ec29c783a65ee7089ff7f8fbcb8}{\texttt{https://github.com/KwaiVGI/LivePortrait/commit7fda9}}} of LivePortrait with default settings, as it produces more stable videos in our experiments. 

For AI-assisted stroke triage, we use a stand-alone video module of \textit{DeepStroke}, without multi-modal fusion or adversarial training. The backbone is a ResNet-34 pretrained on the FairFace dataset \cite{karkkainen2021fairface}. The model parameters are frozen except for the last residual block and the output fully connected layer. 
The cross-validation experiments have a common manual seed to ensure the same fold-wise training/validation data points. 

\begin{figure}[t]
    \centering
\includegraphics[width=\linewidth]{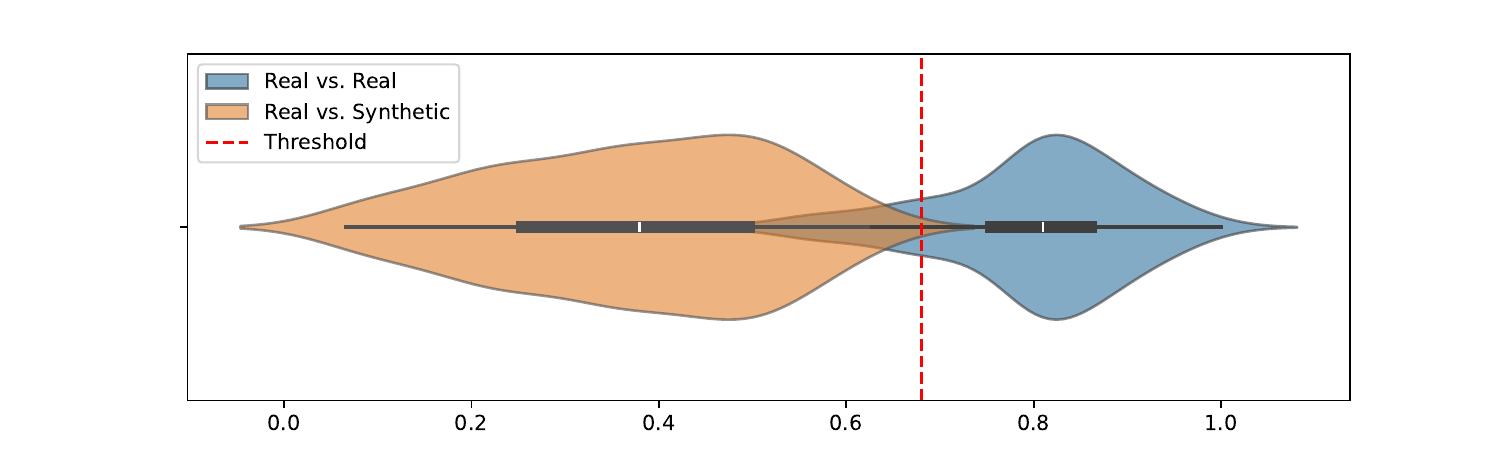}
    \caption{Privacy protection evaluation results. The distributions of facial embedding similarity scores are plotted in the violin plot. \textit{Real vs. Real} refers to scores between two random frames that come from the same real video, and \textit{Real vs. Synthetic} refers to scores between two random frames, one from a real patient video and the other from the corresponding synthetic video. The outlines of the violin shapes are estimated distributions, and a thicker band indicates a higher density. The thicker horizontal lines indicate quarter quantile ranges. The red dashed line at 0.68 is the default face verification threshold in the VGG-Face model~\cite{serengil2020lightface}. 
    }
    \label{fig:privacy}
\end{figure}

\begin{figure}[t]
\centering
\includegraphics[width=\linewidth]{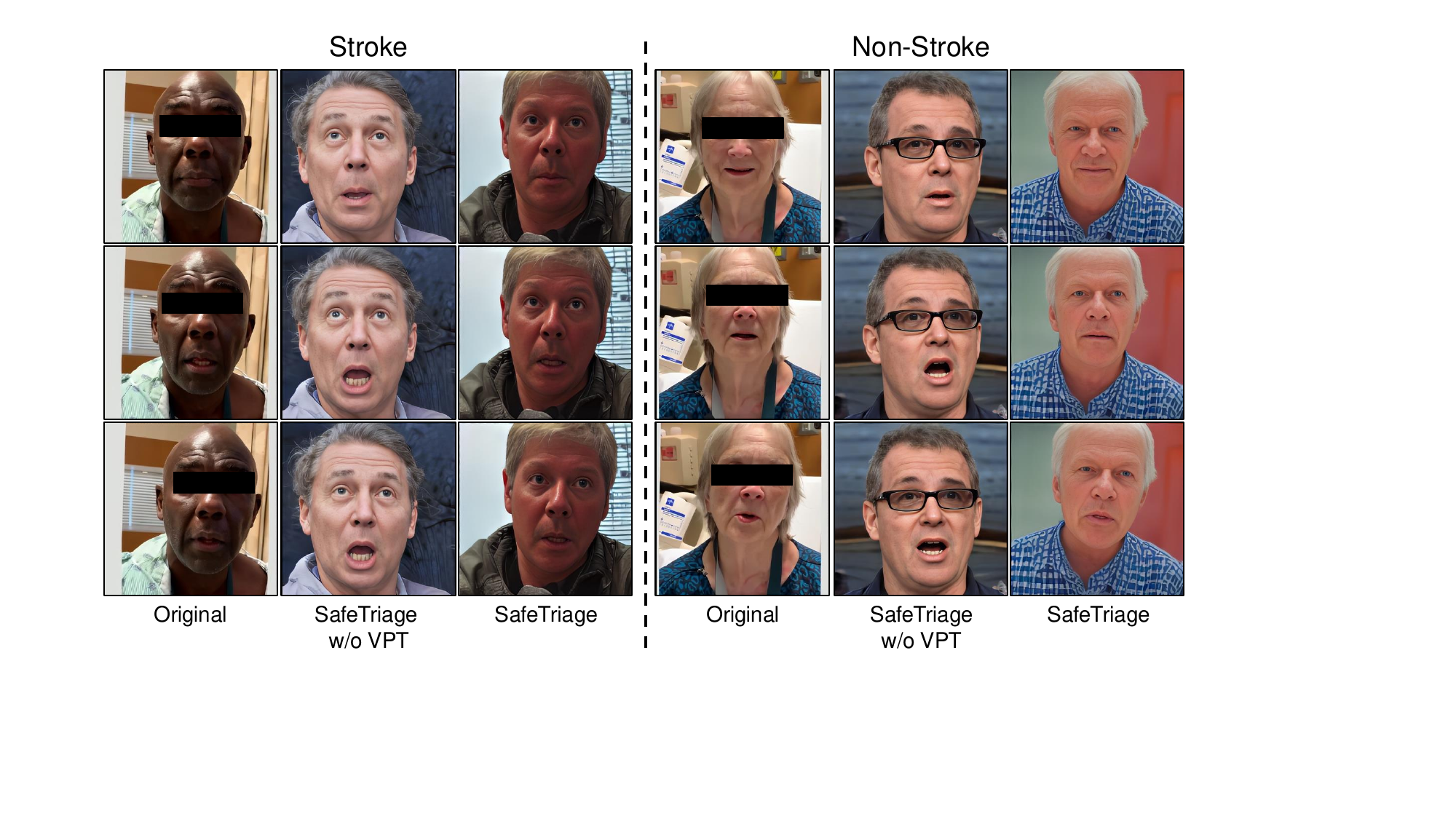}
    \caption{Qualitative ablation study demonstrating the effectiveness of the visual prompt tuning (VPT) module in SafeTriage. The left block shows SafeTriage without vs. with VPT comparison results for a patient with a stroke. The right block shows the comparison for a patient without a stroke. In each block, the columns display the 1st, 200th, and 400th frames at a resolution of $512 \times 512$.}
    \label{fig:ablation}
\end{figure}

\subsection{Result Analysis}
\hspace{\parindent}\textbf{Results of Human Evaluation.} Four independent raters with computer vision expertise participated in evaluating the visual realism of 113 synthetic videos. The overall mean quality score is 2.55 with a standard deviation of 0.65, indicating very good to excellent visual realism since the highest possible score is 3.0. The raters achieved a Fleiss's kappa of 0.608. Among the 113 synthetic videos, 38 are rated as ``Very realistic'' (with score 3.0) by all raters and subsequently selected for further clinician review. A qualified clinician assessed the diagnostic patterns in these selected videos. The mean evaluation score is 0.76, indicating that the majority of retargeted synthetic videos preserved the diagnostic patterns present in the original videos.
Some examples of generated videos are also provided in our GitHub repository. \footnote{\url{https://github.com/Wenchao-M/SafeTriage-Supplementary}}

\textbf{Results of Privacy Evaluation.} The result of privacy protection evaluation is shown in Fig.~\ref{fig:privacy}. The violin plot illustrates the similarity score distribution of two random frames. The pairs are either from the same real video or consist of one from a real video and the other from the corresponding synthetic video. The similarity scores for the pairs from the same real videos are very high overall since they are essentially frames of the same person. When applying SafeTriage and calculating the similarity scores for the real-synthetic pairs, the scores drop to the below 0.6 range, indicating that the synthetic video represents a different identity from the corresponding real video. This result confirms that the generated synthetic videos have minimal privacy leakage, even though some facial features, such as landmarks and edges, are used as conditions for the visual prompt tuning model.

\textbf{Results of AI Stroke Triage.}
Table~\ref{tab:5f} shows the stroke triage results of SafeTriage-generated synthetic videos using the \textit{DeepStroke} AI triage model. Note that the evaluation of the proposed SafeTriage framework involves the performance of the \textit{DeepStroke} video-only module under several train-test schemes: 1. trained and tested with generated videos (Syn-Syn), 2. trained with generated videos and tested with real videos (Syn-Real), and 3. trained with real videos and tested with generated videos (Real-Syn). Meanwhile, the baseline refers to the model's performance when both training and testing are conducted using real videos (Real-Real). To measure the level of agreement in class prediction probabilities, we also report the mean-squared error (MSE) between the probability score vector from the real video baseline and those of other train-test schemes involving synthetic videos.
Furthermore, to demonstrate the effectiveness of our proposed visual prompt tuning (VPT) method, we conduct an ablation study by generating another set of synthetic videos using synthetic faces from an unconditional model (StyleGAN3-FFHQ)~\cite{karras2021alias}, referring to it as the w/o VPT model.

\begin{table}[t]
\centering
\caption{Stroke triage results. 
\textit{Acc.}, \textit{Spec.}, and \textit{Sens.}, represent accuracy, specificity, and sensitivity, respectively. Triage refers to human triage performance, which is a general statistic from the ER department of the data provider. Baseline, w/o VPT, and SafeTriage (ours) results are from the DeepStroke AI triage model. The AI triage performance statistics are adjusted so that the sensitivity level roughly matches with that of human triage.}
\label{tab:5f}
\setlength{\tabcolsep}{4pt}
\resizebox{\textwidth}{!}{%
\begin{tabular}{l|c|c|c|c|c|c|c|c}
\toprule
\textbf{Method}                      & \textbf{Train} & \textbf{Test} & \textbf{Acc.}(\%)$\uparrow$  & \textbf{Spec.}(\%)$\uparrow$ & \textbf{Sens.}(\%)$\uparrow$ & \textbf{F1}$\uparrow$     & \textbf{AUC}$\uparrow$    & \textbf{MSE}$\downarrow$    \\
\midrule
Triage                      & -     & -    & 64.03 & 45.71 & 70.19 & -      & -      & -      \\
\midrule
Baseline                    & Real  & Real & 62.10 & 50.44 & 71.10 & 0.6869 & 0.6885 & 0     \\
\midrule
\multirow{3}{*}{w/o VPT}    & Syn   & Real & 61.83 & 49.33 & 71.10 & 0.6741 & 0.6730 & 0.1386 \\
& Real  & Syn  & 55.62 & 35.11 & 70.77 & 0.5708 & 0.5526 & 0.1358 \\              
                            & Syn   & Syn  & 54.04 & 28.44 & 71.65 & 0.6241 & 0.5883 & 0.1479 \\
                            \midrule
\multirow{3}{*}{\begin{tabular}[c]{@{}l@{}}SafeTriage\\ (Ours)\end{tabular}} & Syn   & Real & 62.06 & 49.78 & 71.21 & 0.6757 & 0.6800 & 0.0875 \\
& Real  & Syn  & 61.06 & 46.81 & 71.21 & 0.6811 & 0.6421 & 0.1058 \\
                            
                            & Syn   & Syn  & 54.98 & 32.22 & 71.43 & 0.6443 & 0.5869 & 0.1017
                            \\
                            \bottomrule
\end{tabular}%
}
\end{table}

The evaluation schemes are designed to assess the potential use cases of the SafeTriage framework. We anticipate that the most common use case is when synthetic videos are used to train AI models, which are subsequently applied to real patient diagnosis, corresponding to the Syn-Real scheme. As shown in Table~\ref{tab:5f}, SafeTriage achieves performance very close to the baseline, with or without VPT. The full SafeTriage framework with VPT gives results that are the closest to the baseline in terms of classification metrics and MSE. Besides the quantitative ablation study regarding the visual prompt tuning model, we also present in Fig.~\ref{fig:ablation} a qualitative comparison of results with vs. without VPT.

Another potential use case can involve using synthetic videos for privacy-preserving diagnosis, where the diagnosis model is trained with videos of real patients who have already consented to data usage. This corresponds to the Real-Syn scheme. This scheme, as shown in the Table~\ref{tab:5f}, results in a slightly lower performance compared to the baseline. We contemplate that this performance drop could be attributed to the more diverse backgrounds in synthetic videos, whereas the real videos have more uniform hospital-setting backgrounds.

The Syn-Syn scheme, in which the AI triage model is both trained and tested on synthetic videos, showed a significant performance drop. One possible reason is that artifacts in the synthetic videos, resulting from imperfections in the conditional generative model and video transfer model, may have amplified effects in the same-domain (Syn-Syn) scheme compared to the diminished effects in cross-domain models (Syn-Real and Real-Syn). 
Future efforts could explore potential solutions to this issue. 

\section{Conclusion and Discussion}
In this study, we have introduced SafeTriage, a novel facial video de-identification framework designed for privacy-preserving stroke triage. SafeTriage generates synthetic videos that anonymize patient identities while retaining crucial diagnostic motion patterns. 
By integrating visual prompt tuning and a large pretrained video motion transfer model, our approach enables zero-shot, privacy-preserving video synthesis.
Through comprehensive evaluation, SafeTriage has been shown to effectively protect patient privacy while maintaining clinical utility. 
We believe that SafeTriage represents a major step forward in addressing challenges associated with sharing privacy-sensitive clinical facial video data, laying the groundwork for enabling the training of large-scale AI foundation models for screening and diagnosing neurological conditions, without compromising patient privacy.

Despite its contributions, SafeTriage has certain limitations. 
First, its generation quality is inherently constrained by the capabilities of the pretrained VMT model. Since our framework is general in terms of which VMT model to use, we expect its performance to improve as more powerful VMT foundation models become available in the future. 
Second, the current SafeTriage framework focuses solely on visual data. However, the value of multi-modal diagnosis, incorporating corresponding audio data, is well-established. 
Future work will integrate audio de-identification to enable the secure sharing of both video and audio data. Additionally, due to time constraints, our clinical evaluation is limited to diagnostic pattern assessment in synthetic videos. In the future, we will conduct broader clinical studies to assess their real-world applicability. Finally, we will explore improved privacy evaluation metrics, such as video-based similarity measures, to better assess identity preservation across frames.

\begin{credits}
\subsubsection{\ackname} Research reported in this publication was supported in part by the National Institute of Neurological Disorders and Stroke of the National Institutes of Health under award number R01NS140292, the T.T. and W.F. Chao Foundation, and the John S. Dunn Foundation. Additionally, part of the experiments were supported by computing resources from the iTiger GPU cluster\footnote{\url{https://itiger-cluster.github.io/}}, funded by NSF award CNS-2318210 and partially by generous contributions from the College of Arts and Sciences and Information Technology Services at the University of Memphis.
\end{credits}

\bibliographystyle{splncs04}
\bibliography{mybib}

\end{document}